\documentclass[pdflatex,sn-mathphys]{sn-jnl}

\jyear{2021}%

\theoremstyle{thmstyleone}%
\theoremstyle{thmstyletwo}%

\theoremstyle{thmstylethree}%

\begin{document}

\title[A Pipeline for \textit{BI} and \textit{RCA} on categorical data]{A Pipeline for Business Intelligence and Data-Driven Root Cause Analysis on Categorical Data}


\author*[1]{\fnm{Shubham} \sur{Thakar}}\email{shubham.thakar@spit.ac.in}

\author[1]{\fnm{Dhananjay} \sur{Kalbande}}\email{drkalbande@spit.ac.in}

\affil*[1]{\orgdiv{Department of Computer Engineering}, \orgname{Sardar Patel Institute of Technology},  \state{Mumbai}, \country{India}}


\abstract{Business intelligence (\textit{BI}) is any knowledge derived
from existing data that may be strategically applied within a
business. Data mining is a technique or method for extracting \textit{BI}
from data using statistical data modeling. Finding relationships
or correlations between the various data items that have been
collected can be used to boost business performance or at the
very least better comprehend what’s going on. Root cause analysis
(\textit{RCA}) is discovering the root causes of problems or events to
identify appropriate solutions. \textit{RCA} can show why an event
occurred and this can help in avoiding occurrences of an issue
in the future. This paper proposes a new clustering + association
rule mining pipeline for getting business insights from data. The
results of this pipeline are in the form of association rules having
consequents, antecedents, and various metrics to evaluate these
rules. The results of this pipeline can help in anchoring important
business decisions and can also be used by data scientists for
updating existing models or while developing new ones. The
occurrence of any event is explained by its antecedents in the
generated rules. Hence this output can also help in data-driven
root cause analysis.}




\maketitle

\section{Introduction}
\textit{BI} is a broad term that comprises data mining, process analysis, and descriptive analytics. The main aim of \textit{BI} is to utilize all the generated data and present informative reports which can anchor business decisions. In this paper, we propose an approach that can be used to extract business insights from data. The results of this pipeline are in the form of association rules which can also be used for root cause analysis or understanding why a particular event occurred. 
We construct a graph \textit G, considering each row to be a node, and connect two nodes if the cosine similarity between them is larger than a threshold.
 We run the hierarchical Louvain’s algorithm \cite{1} on graph \textit G to detect mutually exclusive communities within \textit G and choose the best communities based on their strength and the number of nodes present in each community. The strength of each community is simply estimated as a ratio of the sum of weights of intra-community edges to the sum of weights of the total number of edges in a community. 
Then we apply the apriori algorithm to best communities to mine frequent item sets, association rules, and various metrics such as support, confidence, and lift which are used to evaluate the rules.
Typically for a large dataset, the number of association rules generated is in the order of millions. It becomes difficult to get business insights from these raw rules since they consist of high redundancy. Hence we process them using the Consequent Based Association Rule Summarization technique to summarize the rules based on common consequents. This technique can help us summarize millions of rules into a few rule summaries. \par
The paper is divided into the following sections. A literature survey undertaken about various existing techniques for clustering, association rule mining, and summarizing association rules is described in section 2. Section 3 of this paper explains, in brief, the Louvain algorithm and association rule mining. In Section 4 the proposed method for extracting business insights from data is discussed. Section 5 describes the dataset on which the proposed method is applied and section 6 discusses the results obtained. Section 7 concludes the paper.

\section{Literature Survey}
This section refers to the existing research in the field of root cause analysis and data mining. Various techniques used by researchers for root cause analysis have been discussed here. Researchers have also worked on various techniques for extracting insights via data mining, such approaches have also been discussed below.

\subsection{Community detection}
There are multiple ways to cluster categorical data, one of them is using the Kmodes algorithm. In \cite{2}, researchers discuss the issues with the Kmodes algorithm and propose a community detection-based clustering approach. They have compared the clustering results of Kmodes and community detection and concluded that community detection gives better results. The approach suggested in this paper requires the user to input the number of clusters. In the proposed paper, we have used a modification of this approach that does not require any assumptions on the data/graph.

\subsection{Data to graph}
In \cite{3}, the authors suggest various neighborhood-based methods as well as minimum spanning tree-based methods for converting a dataset to graph form. Some of these methods include the epsilon ball method, k-nearest neighbors and continuous k nearest neighbors. We have used the widely used epsilon ball method for converting the dataset to a graphical form.

\subsection{Business insights via data mining}
The issue with association rule mining is that it produces a huge number of redundant rules which make it nearly impossible to gain insights from those raw rules. In order to tackle this issue, we can reduce the number of produced rules by summarizing them. For example, in \cite{4} the authors discuss that the issue is not in the high number of rules that data mining techniques generate but rather it is in the way we summarize and present these rules. The author further presents the idea to organize the rules in a multilevel format and then summarize them to eliminate redundant rules. Other approaches include clustering similar association rules and using the concept of rule cover to prune association rules \cite{5}. In order to identify unique and essential association rules some researchers have devised a representative basis and they also present an efficient way of computing this representative basis \cite{6}. In \cite{7}, the researchers discuss the drawbacks of association rule mining and propose a simple but effective consequent-based association rules summarization technique that has been used in this paper to filter out redundant rules. \par
Another method for extracting insights from categorical data and understanding relations between columns is to convert the categorical variables to continuous variables using ordinal encoding \cite{16} and then using a heatmap \cite{11} to find relations between the columns. However, this method cannot be applied for columns that are not ordinal. Our proposed method solves this issue since it can be applied to ordinal as well as nominal columns.

\subsection{Data Driven Root Cause Analysis}
In \cite{12} the authors have pointed out that the state-of-art work on \textit{RCA} suffers from interpretability, accuracy, and scalability. They have proposed an approach where an LSTM autoencoder is used to find anomalies and SHAP, which is a game theory based approach is used to find the root cause of the issue in fog computing environment. Along similar lines, authors in \cite{13} have used a process mining based approach for securing IoT systems. In \cite{14} the authors present 2 data driven methods for \textit{RCA}, sequential state switching and artificial anomaly association. These approaches have been developed for distributed system anomalies and have been tested using synthetic data and real world data based on the Tennessee Eastman process (TEP). All these data driven \textit{RCA} and mining approaches are specific to a particular domain and lack generality. The proposed solution aims to provide a general technique that can be applied to any categorical dataset.

\section{Preliminaries}

This section gives a brief introduction to Louvain's community detection algorithm and why was it preferred over Kmodes. It also explains association rule mining in brief.

\subsection{Louvains Community detection}

The Louvain method \cite{1}, a hierarchical clustering method that was first published in 2008, has become a popular choice for community detection. It is a bottom-up folding technique that optimizes modularity \cite{15}. The method is divided into passes, each of which consists of the repetition of two phases. Every node is first given a unique community. Therefore, in the initial phase, there will be an equal number of communities as nodes. The algorithm then assesses every node's increase in modularity if we move it from its community to that of a neighbor. If there is a positive gain, the node is then assigned to that community; otherwise, it stays in the community to which it was originally assigned. All nodes undergo this procedure consecutively and repeatedly until no more increase in modularity is possible, at which point the first pass is finished.\par

Over the years, other additional clustering techniques have been discovered and suggested (for a recent survey, see \cite{8}). The k-modes technique was created by Huang \cite{9} to cluster categorical data. The k-modes approach is a partitional clustering method and needs the number of clusters as an input. There are methods such as the elbow method which are used to find the optimal number of clusters but typically this method returns a smooth curve making it challenging to detect the elbow. Furthermore, the Kmodes algorithm is quite sensitive to the initial cluster centers selected, and a poor selection may lead to cluster configurations that are very undesirable.\par
The Louvain algorithm does not require any assumptions on the data/graph and is deterministic in contrast to conventional Kmodes with random initialization. Moreover, it is faster when compared to its peers and runs in \textit{O(n.log(n))}. After considering these factors we have used the Louvain community detection algorithm in the proposed pipeline.

\subsection{Association rule mining}
Association rule mining, a data mining technique helps unearth intriguing relationships in data \cite{17}. Market basket analysis \cite{18}, which identifies products that are typically bought together in a supermarket, is a problem where association rule mining is often applied. For example, the rule \{eggs, butter\}\textrightarrow \{bread\} found in the sales data of a supermarket would indicate that if a customer buys eggs and butter together, they are likely to also buy bread. The association rule generation is split into 2 steps:
\begin{enumerate}
  \item A minimum support threshold needs to be provided based on which frequent itemsets will be filtered.
  \item To generate rules from the frequent item sets a minimum confidence threshold needs to be provided.
\end{enumerate}

For a given rule X\textrightarrow Y,
support is nothing but the frequency of X and Y coming together in a transaction. Confidence is the conditional probability of getting Y given that X is already present in the transaction. The issue with confidence is that if the support for Y in the dataset is high then all rules having Y as consequent will have high confidence, hence we use another metric called lift which shows whether X and Y are positively correlated (given X probability of Y being present increases) or negatively correlated. X and Y are said to be positively correlated if the value of lift is greater than 1.


\begin{figure}[h]%
\centering
\includegraphics[width=0.7\paperwidth]{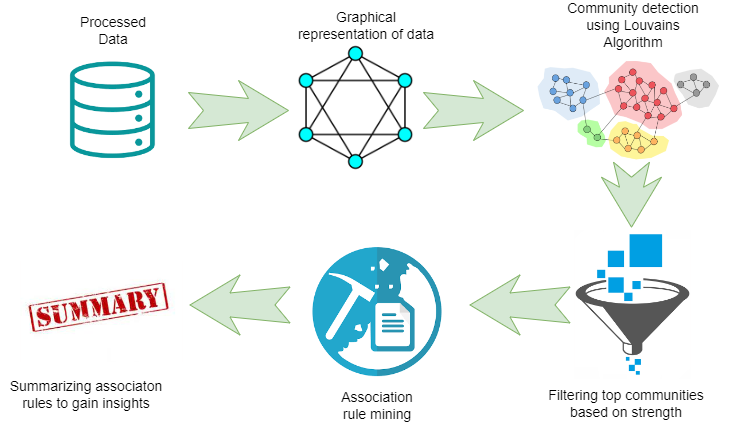}
\caption{Methodology}\label{fig1}
\end{figure}

\section{Methodoloy}

This section explains the proposed method which consists of 3 steps including converting data to graph, modularity-based community detection, association rule mining, and summarization. Figure 1 explains the steps involved in the method.

\subsection{Graph conversion}
The proposed method uses the epsilon ball method for converting the dataset to a graph \textit{G}. We assume each row to be a node and draw a weighted edge between 2 nodes if the cosine similarity between the 2 rows is greater than epsilon.

\begin{figure}[h]
\centering
\includegraphics[width=0.6\paperwidth]{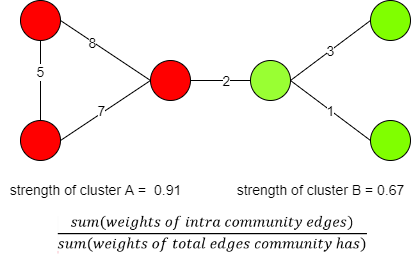}
\caption{Strength Calculation}
\end{figure}

\begin{figure}[h]
\centering
\includegraphics[width=0.6\textwidth]{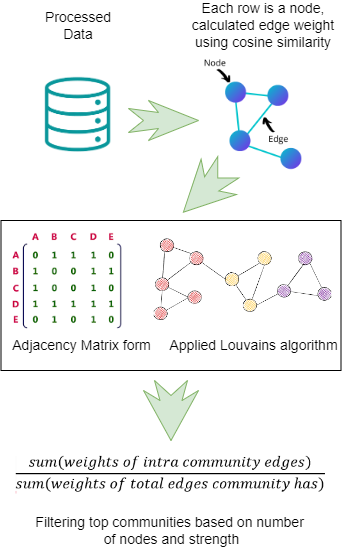}
\caption{Graph conversion and community detection}
\end{figure}

\subsection{Modularity-based community detection}
On graph \textit G, we apply the Louvain technique \cite{1} to find strongly related groups of nodes. The best communities based on strength and size will be filtered. The strength of a community is calculated using the formula stated in figure 2. The communities having a higher number of intra-community edges with more weight will generally have higher strength. Figure 2 depicts an example of strength calculation. Note that modularity \cite{15} is a measure for the complete graph and not individual clusters, hence we define a new measure to compare the strength of clusters. Figure 3 explains the steps involved in detecting top clusters for applying association rule mining.

\subsection{Association Rule Mining and summarization}
One issue with association rule mining is that it generates too many rules since it considers all permutations and combinations. Many of these generated rules carry the same information and are redundant. Generally, the number of essential rules is very few when compared to the number of redundant rules, which makes it difficult to gain insights from these raw rules. We further process these rules using a consequent-based association rule summarization (CARS) \cite{7} technique to filter essential rules. \par
The CARS \cite{7} technique follows 3 steps:
\begin{enumerate}
  \item Filter rules having only 1 consequent.
  \item Perform a group by operation on the consequents so that we only have unique consequents and rank the antecedents based on their frequency of occurrence.
  \item Display the minimum and maximum of each interestingness measure with the rule summary. 
  \end{enumerate}
Table 1 showcases an example of how to apply the CARS \cite{7} technique on raw rules.\par
Using this method, a huge number of association rules are reduced to a small number of rule summaries, each of which only carries one unique consequent. Firstly, since a majority of business users assess a rule's usefulness by assessing the relevance of its consequent, it makes sense to have only 1 consequent. This makes it easier to focus on rules with relevant consequents. Secondly, the antecedents are ranked in each rule summary based on their correlation with the consequent, e.g. in the rule summary in table 1, A is more closely related to C than B since the frequency count associated with A is higher. This helps in determining which antecedents are more closely related to the consequent. Lastly, each rule summary also has a range of interestingness measures which can again help in evaluating the rule summary.

\begin{table}[h!]
  \begin{center}
  \begin{minipage}{250pt}
    \caption{Association Rule Summarization}
    \label{tab:table1}
    \begin{tabular}{|l|l|l|l|l|}
    \hline
    Antecedent & Consequent & Support & Confidence & Lift \\ \hline
    (A, B) & C & 20\% & 50\% & 1.5  \\ \hline
    (A) & C & 30\% & 70\%  & 2.4 \\ \hline
    ((A, 2), (B, 1)) & C & [20, 30]\% & [50, 70]\% & [1.5, 2.4]\\ \hline
    \end{tabular}
    \end{minipage}
  \end{center}
\end{table}

\section{Dataset}
We have tested our method on a private skin ointment dataset that contains both medical and beauty products. Some important columns include category, manufacturer name, country of origin, area of application, rating(numeric), price(numeric), is prescription needed(bool), side effects(bool), country of sale, male/female, marketer name, skin type, target disease, texture. The processed data has 14 columns and around 10 thousand rows. We converted the 2 numeric attributes to categorical ones using suitable intervals. Further, the dataset was converted to a graphical form using the epsilon ball method.

\section{Results}
Applying Louvain's algorithm to the data resulted in 3 clusters. Out of the 3 clusters, rule mining was applied on cluster 2 since it had higher strength and a good number of nodes. With a support threshold of 20\% and a confidence threshold of 50\% more than a million rules were generated. These rules were summarized using the CARS \cite{7} technique to 25 rule summarizes. The most important rule summaries are presented in table 2 and table 3.

\begin{table}[h!]
  \begin{center}
  \begin{minipage}{250pt}
    \caption{Extracting important features}
    \label{tab:table1}
    \begin{tabular}{|l|l|l|l|l|}
    \hline
    Antecedents &  &  & & Consequent\\ \hline
    manufac name & country & category & area of app & rating\\ \hline
    Lifeline & USA & Beauty & scalp & 4-5\\ \hline
    Garnier & France & Beauty & face & 3-4\\ \hline
    J\&J & USA & Medical & body & 2-3\\ \hline
    \end{tabular}
    \end{minipage}
  \end{center}
\end{table}

Table 2, represents all rule summaries having ratings as consequents, and top antecedents for those consequents. These results showcase that if a data scientist wants to train a model to predict the rating then manufacturer name, country of origin, category, and area of application are top features that need to be considered.

\begin{table}[h!]
  \begin{center}
  \begin{minipage}{250pt}
    \caption{Root cause analysis}
    \label{tab:table1}
    \begin{tabular}{|l|l|l|l|l|}
    \hline
    Antecedents &  &  & & Consequent\\ \hline
    category & manufac name & area of app & country & price(usd)\\ \hline
    beauty & Lifeline & face & USA & 200-max\\ \hline
    \end{tabular}
    \end{minipage}
  \end{center}
\end{table}

Table 3, represents a rule summary that has the highest price category as a consequent. This summary can help in understanding the root cause behind a skin product being in the highest price category. It suggests that if a product is of beauty category, was manufactured by Lifeline, is applied on the face and the country of origin is the USA then it will be costly.

\section{Conclusion}
This paper has presented a method for effective root cause analysis and business insights generation using community detection-based clustering and association rule mining. This paper uses the epsilon ball method for converting data to a graph format, however other methods such as knn, cknn, etc mentioned here can be explored. Louvain community detection is preferred over Kmodes as it is a hierarchical clustering method and allows finding clusters without any assumptions on the data/graph. Since the apriori algorithm has a high time complexity, clustering helps in filtering out data points and thus reducing the time taken by apriori. Moreover, clustering also brings similar data points together, this helps in getting essential rules even at higher levels of support threshold. Association rule mining is then applied on top clusters and the results are summarized using the CARS \cite{7} summarization techniques. These summarized rules can help in root cause analysis, business insights generation, and also understanding of the correlation between various features of the categorical dataset.

\end{document}